\UseRawInputEncoding

\documentclass[journal]{IEEEtran}
\ifCLASSINFOpdf
\else
\fi

\usepackage{cite}
\usepackage{graphicx}
\usepackage[ruled,vlined]{algorithm2e}
\usepackage{pifont}
\usepackage{booktabs}
\usepackage{amssymb}
\usepackage{amsmath}
\usepackage{color}
\usepackage{multirow}
\usepackage[table,xcdraw]{xcolor}
\usepackage{stfloats}

\begin{document}
%

\title{Dual-modal Prior Semantic Guided Infrared and Visible Image Fusion for Intelligent Transportation System}

%

\author{Jing~Li, Lu~Bai~\IEEEmembership{IEEE~Member}, Bin~Yang, Chang~Li, Lingfei~Ma, Lixin~Cui~\IEEEmembership{IEEE~Member}, and~Edwin R. Hancock~\IEEEmembership{IEEE~Fellow}} 

%
%

\markboth{Journal of \LaTeX\ Class Files,~Vol.~x, No.~x, x~x}%
{Shell \MakeLowercase{\textit{et al.}}: Bare Demo of IEEEtran.cls for IEEE Journals}
%



\maketitle

\begin{abstract}
Infrared and visible image fusion (IVF) plays an important role in intelligent transportation system (ITS). The early works predominantly focus on boosting the visual appeal of the fused result, and only several recent approaches have tried to combine the high-level vision task with IVF. However, they prioritize the design of cascaded structure to seek unified suitable features and fit different tasks. Thus, they tend to typically bias toward to reconstructing raw pixels without considering the significance of semantic features. Therefore, we propose a novel prior semantic guided image fusion method based on the dual-modality strategy, improving the performance of IVF in ITS. Specifically, to explore the independent significant semantic of each modality, we first design two parallel semantic segmentation branches with a refined feature adaptive-modulation (RFaM) mechanism. RFaM can perceive the features that are semantically distinct enough in each semantic segmentation branch. Then, two pilot experiments based on the two branches are conducted to capture the significant prior semantic of two images, which then is applied to guide the fusion task in the integration of semantic segmentation branches and fusion branches. In addition, to aggregate both high-level semantics and impressive visual effects, we further investigate the frequency response of the prior semantics, and propose a multi-level representation-adaptive fusion (MRaF) module to explicitly integrate the low-frequent prior semantic with the high-frequent details. Extensive experiments on two public datasets demonstrate the superiority of our method over the state-of-the-art image fusion approaches, in terms of either the visual appeal or the high-level semantics.
\end{abstract}

\begin{IEEEkeywords}
 Infrared image, Visible image, Image fusion, High-level vision task.
\end{IEEEkeywords}

%
\IEEEpeerreviewmaketitle

\begin{figure}[!t]
\centering
\includegraphics[width=3.5 in]{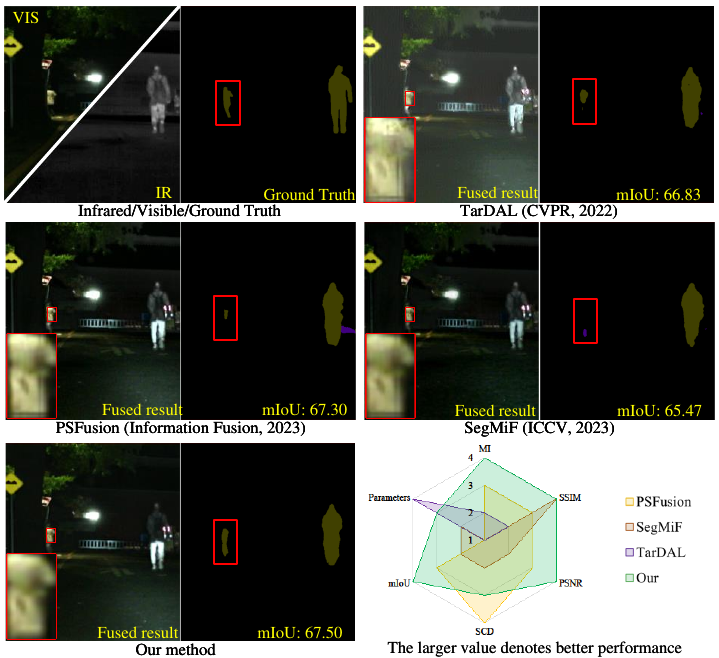}
\caption{Comparisons of our method and several task-driven methods in high-level vision task and image fusion task. Our method has the better performance both in qualitative and quantitative experiments of the two tasks, and our method also has less parameters than other semantic segmentation task-driven methods, such as PSFusion, SegMiF.}
\label{FIG:1}
\end{figure}

\section{Introduction}
\IEEEPARstart{M}{ulti-modality} images can describe the scene from various perspectives, which can aggregate more complementary information for ITS~\cite{zhang2023visible,li2023lrrnet,karim2022current,ma2019infrared,pei2023calibnet}. Therefore, the multi-modality image fusion has been a popular technology to address the data biasing problems that are caused by the single sensor in ITS~\cite{ju2022ivf,li2023sosmaskfuse,zhang2021abmdrnet}. Generally speaking, the infrared and visible sensors have diverse imaging mechanisms. Specifically, the infrared sensor is sensitive to the thermal radiation information, that are important for ITS and can highlight the saliency targets in the nighttime or some harsh environments (e.g., the pedestrian or car)~\cite{xue2022nighttime}. However, the infrared sensor suffers from the loss of details due to the insensitivity of the thermal radiation in the texture space~\cite{tang2022superfusion,zhao2023metafusion}. In contrast, the visible sensor produces images by capturing the reflected light. Thus, the visible images contain more texture details and can provide abundant scenario cues or traffic movements for ITS, e.g., road boundaries, the plate number of the cars, etc. With the multi-modality image fusion to hand, ITS can benefit from both infrared and visible images in harsh environments or complex traffic scenarios, e.g., the object detection or the semantic segmentation.

\begin{figure}[!t]
\centering
\includegraphics[width=3.5in]{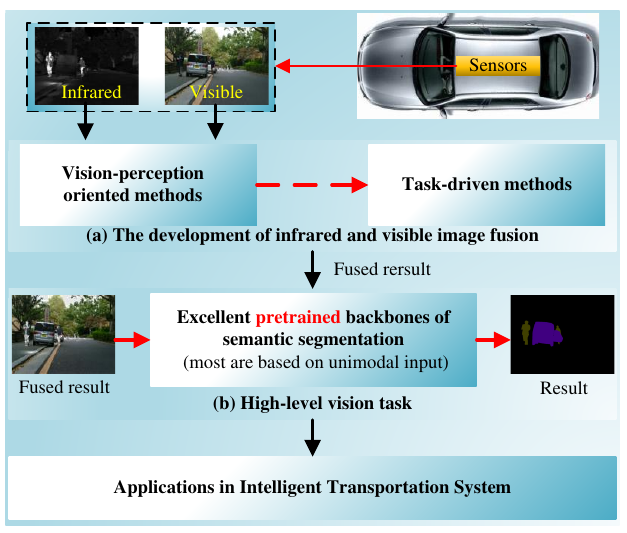}
\caption{The overview and development of infrared and visible image fusion in ITS. We also discuss the differences between our method and recent task-driven methods.}
\label{FIG:2}
\end{figure}

The excellent backbones of the high-level vision tasks are another alternative powerful tools for ITS, and are usually defined based on the unimodal input, e.g., the Segformer~\cite{xie2021segformer}, the ViTadpter~\cite{chen2023vitadapter}, the InternImage~\cite{wang2023internimage}, etc. However, since the single modal input image suffers from the data biasing problem, it provides less meaningful information for high-level vision tasks than multi-modality images. On the other hand, the inputs of multi-modality images for novel backbones usually need tedious re-designs, this may further limit the performance of  high-level vision tasks associated with both infrared and visible images. To solve the contradiction problem in high-level vision tasks of ITS, it is significant to employ the fusion of the infrared and visible images that can produce a single fused result with complementary information as the input for the excellent backbones.

Unfortunately, most of the existing infrared and visible image fusion methods suffer from some common drawbacks. In early works, the traditional methods mainly employ the multi-scale transform~\cite{shreyamsha2015image,burt1987laplacian}, the sparse representation~\cite{zhang2018sparse}, the saliency analysis~\cite{bavirisetti2016two,wang2023interactively}, the subspace transform~\cite{mitianoudis2007pixel}, and other hybrid methods~\cite{liu2015general}, to generate the fusion image. Since these traditional methods need hand-crafted complicated fusion strategies, and cannot provide an end-to-end learning framework. These traditional methods usually have lower application performance in many down-stream tasks. In recent years, With the successful development of deep learning (DL), many DL-based fusion methods are proposed to solve the issues of the traditional methods. Specifically, these DL-based methods can fuse the images through an end-to-end way, associated with convolutional neural networks (CNNs)~\cite{li2021rfn,zhang2020ifcnn}, generative adversarial networks (GANs)~\cite{ma2019fusiongan,li2020attentionfgan,ma2020ddcgan,li2020infrared,li2020multigrained}, the transformer~\cite{vs2021image,li2022cgtf,ma2022swinfusion,wang2022swinfuse,li2023tfiv}, or other DL models~\cite{zhao2023ddfm,zhao2023cddfuse}. Thus, these recent DL-based method can generate satisfactory fused results, significantly improving the application performance. However, either the traditional methods or the DL-based methods focus more on the reconstructing raw pixels rather than the applications in high-level vision tasks, i.e., these existing methods tend to unilaterally pursuing the visual appeal. This drawback limits their performance in TIS. 

\begin{figure*}[!]
\centering
\includegraphics[width=6 in]{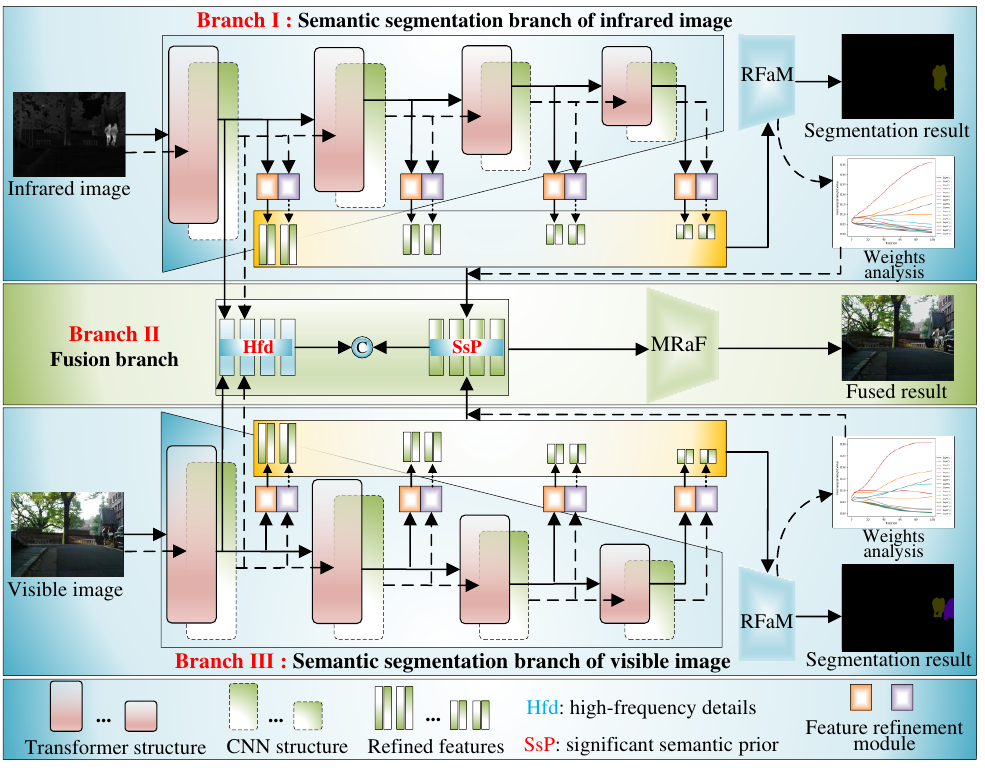}
\caption{The overall framework of the proposed method based on dual-modality semantic guided image fusion strategy for high-level vision tasks, which includes two parallel semantic segmentation branches with refined feature adaptive-modulation (RFaM) module and multi-level representation-adaptive fusion (MRaF) module. More details of weights analysis are shown in Fig. 5.}
\label{FIG:3}
\end{figure*}

To address the above challenges and improve the performance of the fused results in high-level vision tasks, the task-driven fusion methods are proposed~\cite{tang2023rethinking,liu2023SegMiF,liu2022target,tang2022seafusion}, typical instances include PSFusion~\cite{tang2023rethinking}, SegMiF~\cite{liu2023SegMiF} and TarDAL~\cite{liu2022target}. Specially, these methods tend to employ the cascaded structure to constrain the fusion network with a down-stream vision task, and then adopt the deeper features of the feature extraction module to insert the semantic information into fusion results. However, these methods still suffer from the following shortcomings. \textbf{(a)} Their fusion and down-stream vision tasks are deeply coupled with the cascaded structure to seek the unified suitable features, and simultaneously fit different tasks~\cite{liu2023SegMiF,tang2022seafusion}, resulting in the performance degradation of the semantic information guidance. \textbf{(b)} They neglect the diverse domain variation of infrared and visible images, and only employ the single semantic segmentation network to insert dual-modality semantic cues into fusion tasks~\cite{tang2023rethinking,liu2023SegMiF}, leading to the confusion of the independent modal semantic perception of infrared or visible images. \textbf{(c)} They fail to perceive the significant semantic features~\cite{tang2023rethinking,liu2023SegMiF,tang2022seafusion,liu2022target}, and only employ the pretrained semantic segmentation model to constrain the fusion task by the semantic loss or insert the redundancy semantic information into the fusion task.

Motivated by the above findings, in this work we propose a novel dual-modal prior semantic guided image fusion method for high-level vision tasks. To avoid the unified suitable features for different tasks, we design two parallel semantic segmentation branches rather than one single cascaded structure to extract the independent semantics of infrared and visible images. Since the parallel semantic segmentation branches can provide dual-modality semantics by considering the diverse domain variation of infrared and visible images, the proposed method avoids the confusion of the single semantic perception for two images. Besides, to reduce the redundancy semantic information, we propose to insert RFaM into the two semantic segmentation branches. We conduct two pilot experiments based on the two branches to seek the significant prior semantic of infrared and visible images. Based on the pilot experiment, we adopt the dual-modal prior semantics to guide the fusion task by simultaneously training the two semantic segmentation branches and the fusion branch. Specifically, the semantic segmentation branches aim to provide significant prior semantics for the fusion branch in the training process. Furthermore, to achieve the better performance of impressive visual effect, we further investigate the frequency response of the significant prior semantic features, and then we introduce MRaF to combine high-frequency details with low-frequency significant semantic features. 

Fig. 1 shows that the proposed method achieves better performance both on the fusion task and the semantic segmentation task than the previous task-driven methods. Specifically, we select a typical region from every fused result and zoom it in the left bottom of the fused images, which shows that our fused result contains clearer details and smoother contour than other methods, and has better performance of quantitative experiments in most metrics. Besides, the semantic segmentation experiments demonstrate that our method has the better segmentation result with the largest mIoU.

Overall, the contributions of this paper are summarized as follows:
\begin{itemize}
\item To improve the quality of the learned representation for the fusion result both in the high-level and fusion tasks, we refine and explore the infrared and visible features that are semantically distinct enough to guide the fusion model.

\item To explore the dual-modal significant prior semantic information independently, we employ the refined feature adaptive-modulation (RFaM) mechanism in two parallel semantic segmentation branches, aiming to extract the significant semantic features.

\item To improve the performance of the image reconstruction, we propose a multi-level representation-adaptive fusion (MRaF) module, which explicitly integrates the low-frequent significant semantic feature with high-frequent details to enhance both the low- and high-level feature representation.

\item Ablation studies show the effectiveness of our fusion strategies, and we compare the proposed method with seven high-level vision task-driven and vision-perception oriented methods on two public datasets, and the experiments demonstrate the superiorities of our method.
\end{itemize}

We organize the rest of this paper as follows: in section \uppercase\expandafter{\romannumeral2}, some related works about the infrared and visible image fusion are presented, including the vision-perception and high-level vision task driven image fusion methods. In section \uppercase\expandafter{\romannumeral3}, we show and analyze the details of our method. Meanwhile, the experiments of the vision-perception and semantic segmentation are presented in section \uppercase\expandafter{\romannumeral4}, followed by some conclusions in section \uppercase\expandafter{\romannumeral5}. 

\section{Related works}
\label{}
The infrared and visible image fusion has been developing from focusing on the visual appeal to the task-driven fusion models. Therefore, we first review the vision-perception oriented IVF methods, which also includes reprehensive IVF methods in ITS, and then the existing typical task-driven methods are discussed. 

\subsection{Vision-perception oriented IVF}
The earlier researches of IVF mainly emphasize on the visual appeal, and the methods can be classified into traditional methods and DL based methods. The former methods mainly take the mathematical transformation to extract image features, which are then fused by the well-designed fusion rules to produce the fused result. According to the different mathematical transformation and fusion rules, the traditional methods include the multi-scale transformer-based methods (MST)~\cite{shreyamsha2015image,burt1987laplacian}, the sparse representation-based methods~\cite{zhang2018sparse}, the saliency-based methods~\cite{bavirisetti2016two}, the subspace-based methods \cite{mitianoudis2007pixel}, and the hybrid methods~\cite{liu2015general}. However, all the traditional methods need to manually design the complicated fusion strategies, influencing their performance.

To address the above issues of the traditional methods, deep learning is introduced into IVF tasks, because it can fuse the infrared and visible image with an end-to-end way. Considering the structures of the DL models, the DL-based fusion methods mainly contain the CNN-based methods, the GAN-based methods, and the Transformer-based methods~\cite{xu2020u2fusion,li2021different}. Specifically, the CNN-based methods design the parallel convolution kernels to capture the shift-invariance and locality of the extracted image features~\cite{li2021rfn,zhang2020ifcnn}. The GAN-based methods fuse the images with an adversarial game, and most of them extract the image features in the generator or the discriminator through the convolution operations~\cite{ma2019fusiongan,ma2020ddcgan,li2020infrared,li2020multigrained,li2020attentionfgan}. Unfortunately, both the CNN- or GAN-based methods focus more on the local features, and thus fail to consider the long-range dependency of the features. To overcome this shortcoming, the transformer is applied into IVF tasks. Since these transformer-based methods integrate the transformer and the CNN \cite{li2022cgtf,vs2021image}, the methods can simultaneously extract both the local features and the long-range dependencies~\cite{ma2022swinfusion,wang2022swinfuse}.

Although the above vision-perception oriented methods can produce satisfactory fused result with good visual effects, they still focus more on reconstructing the raw pixels rather than the applications in high-level vision tasks. This in turn limits their performance on ITS. For example, Ju \textit{et al}. propose an IVF-Net to fuse the infrared and visible images for ITS~\cite{ju2022ivf}, whereas, they still only focus on the vision appeal and thus fail to consider the high-level vision task on ITS.

\subsection{High-level vision task-driven IVF}
In the development of IVF, the vision-perception oriented methods fail to meet the demands of high-level vision tasks, such as ITS. To overcome the shortcoming, several task-driven methods are proposed (see Fig.~2 for the detailed development of IVF). For instance, Tang~\textit{et al}. have proposed an IVF model with an additional pretrained semantic segmentation network SeAFusion~\cite{tang2022seafusion}, which can insert the semantic information to the fused result by semantic loss. Liu \textit{et al}. have combined the target detection with IVF to propose a task-driven fusion methods, termed as TarDAL~\cite{liu2022target}. This method can constrain the fusion task to capture the semantic information from the target detection process. After that, Liu \textit{et al}. have also proposed another task-driven IVF model SegMiF~\cite{liu2023SegMiF}, which employs a cascaded structure to combine the image fusion with the semantic segmentation. Tang \textit{et al}. have adopted the deeper features of the feature extraction branch to introduce the semantic information into the scene restoration branch, and propose a task-driven IVF model PSFusion~\cite{tang2023rethinking}. This method demonstrates the potential and necessary of the image-level fusion compared to the feature-level fusion for high-level vision tasks. 

However, the above mentioned task-driven IVF methods also suffer from several problems. For example, the SegMiF method employs the cascaded structure, and the fusion and high-level vision tasks are deeply coupled to seek unified suitable features and  simultaneously fit different tasks. This in turn results in the performance degradation of the guidance of semantic information. Moreover, the PSFusion and SegMiF methods neglect the diverse domain variation of the infrared and visible images, and employ the single semantic segmentation network to insert the dual-modal semantic cues into the fusion task, leading to the confusion problem of the independent modal semantic perception for infrared or visible images. Finally, the TarDAL, PSFusion and SegMiF methods all fail to perceive the significant semantic features, and just employ the pretrained semantic segmentation model to constrain the fusion task by semantic loss or insert the redundancy semantic information into the fusion task.

To address the drawbacks of the mentioned existing IVF methods, we propose a dual-modal semantic guided image fusion method for high-level vision tasks, which includes two parallel semantic segmentation branches rather than a single cascaded structure to extract the independent semantics of infrared and visible images, respectively. Furthermore, we propose RFaM to refine and explore the significant prior semantic features, which can guide the fusion task by simultaneously training the semantic segmentation branches and the fusion branch. Finally, we also design MRaF modules to combine the high-frequent details with the low-frequent significant semantic features. 

\begin{figure}[!t]
\centering
\includegraphics[width=3.5in]{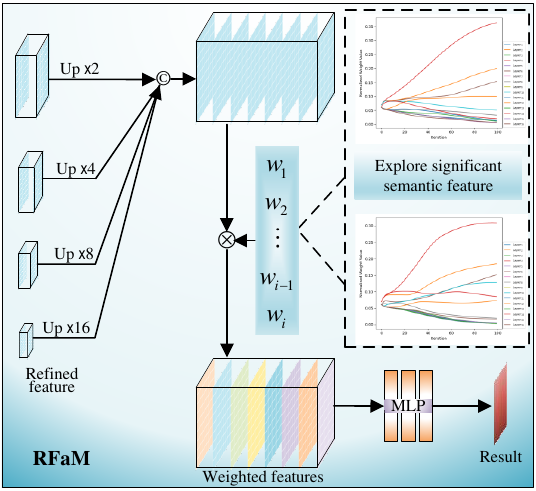}
\caption{The architecture of the refined feature adaptive-modulation (RFaM). More details of weights analysis are shown in Fig. 5.}
\label{FIG:4}
\end{figure}

\section{The proposed method}
\label{}
In this section, we first define the overall framework of the proposed method, and then give the detailed definition of the semantic segmentation branches. Moreover, we introduce the feature refinement module, the feature-adaptive modulation module and the multi-level representation-adaptive fusion module. Finally, we give the details of the loss function.

\subsection{The overall framework}
The purposes of our proposed method simultaneously include two aspects, i.e., \textbf{(a)} the proposed method should provide the visual appealing fused image, and \textbf{(b)} the fused image needs to explore the significant semantic information that fits the requirement of the high-level vision task. Therefore, how to learn the significant semantic feature of each image to guide the image fusion is important for the proposed method. To this end, we propose a dual-modal prior semantic guided image fusion method by designing two parallel semantic segmentation branches, and each branch has the same structure. The overall framework of our proposed method is shown in Fig.~3. To further explore the significant semantic features of each image, we conduct two pilot semantic-segmentation experiments based on the Branch \uppercase\expandafter{\romannumeral1} and the Branch \uppercase\expandafter{\romannumeral3} of Fig.~3. Then, we analyze the weights of the refined feature adaptive-modulation (RFaM) module to perceive the features that are semantically distinct enough in each pilot semantic segmentation experiment. We use the significant prior semantics to guide the image fusion task during the training process. Specifically, in the image fusion task, we integrate the Branch \uppercase\expandafter{\romannumeral2} with the Branch \uppercase\expandafter{\romannumeral1} and the Branch \uppercase\expandafter{\romannumeral3} to reconstruct fused image. In this step, the Branch \uppercase\expandafter{\romannumeral1} and the Branch \uppercase\expandafter{\romannumeral3} can provide significant semantics for the fusion task (i.e., the Branch \uppercase\expandafter{\romannumeral2}). Therefore, our proposed method can capture enough semantic information to meet the requirements of high-level vision tasks. To achieve the impressive visual effect in our fused result, we investigate the frequency response of the significant semantic features, and propose a multi-level representation-adaptive fusion (MRaF) module to further explore the adaptive features that explicitly integrate the low-frequent significant semantic features with
the high-frequent details.

\begin{figure}[!t]
\centering
\includegraphics[width=3.5in]{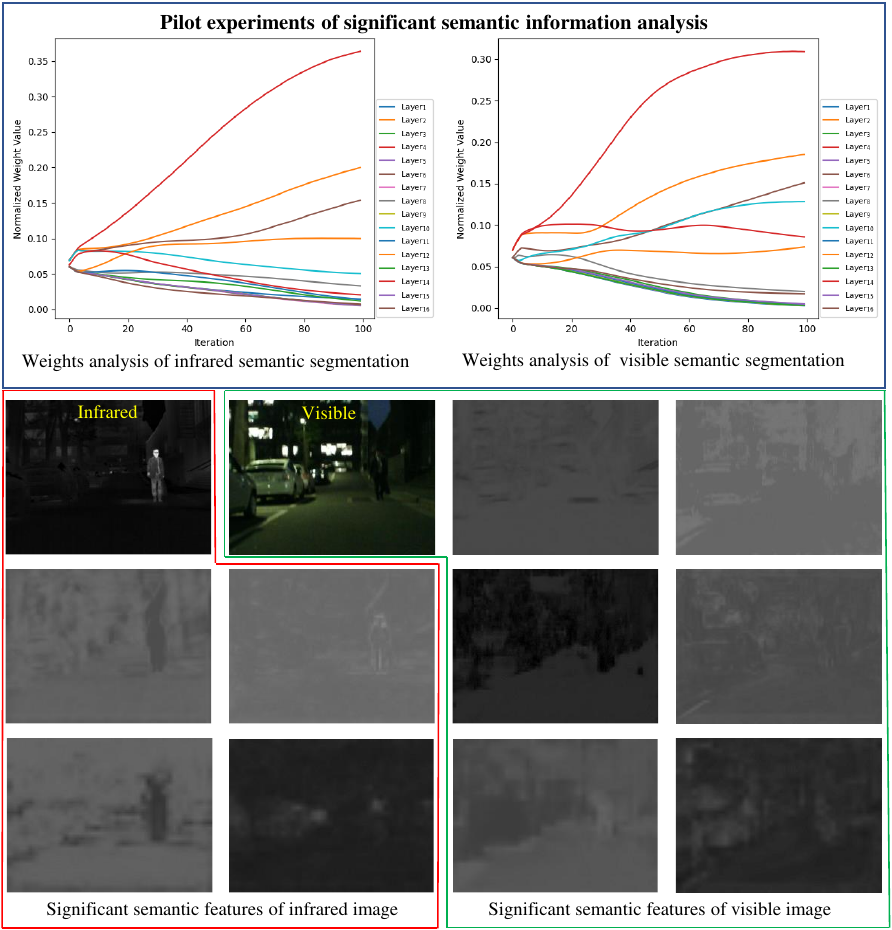}
\caption{Significant semantic information analysis of pilot experiment.}
\label{FIG:5}
\end{figure}

\subsection{Semantic segmentation branches of the infrared and visible images}
The infrared and visible images usually have diverse modalities, and the semantic perception of each independent modal is benefit to the extraction of the significant infrared or visible semantic feature. Thus, in our proposed method, two parallel semantic segmentation branches are designed. For each branch, we proposed to employ the parallel structure to extract the global and local features based on the transformer and the CNN, respectively. Note that, the transformer and CNN have significantly different learning structures. Specifically, the transformer can capture the long-distance relations through the self-attention module and extract the global feature. By contrast, the CNN has the attribute of the translation invariant and the local inductive bias, and is suitable for the extraction of the local feature. Due to the significant difference between the Transformer and the CNN, it is hard to align the intermediate features in terms of the semantic domain~\cite{zhao2023cumulative}. To address this issue, we design a parallel structure based on the transformer and CNN networks in each branch to explore the global-local features, and the architecture is shown in Fig.~3. We refer to the references~\cite{xie2021segformer} and~\cite{ronneberger2015u} to design the structures of the transformer and the CNN respectively, and the detailed definitions are formulated as follows, i.e.,
\begin{equation}
\begin{split}
 \hat{F}_{ir/vis}^{i}=f_{gl}^{i}(\hat{F}_{ir/vis}^{i-1}),
\end{split}
\end{equation}
\begin{equation}
\begin{split}
 \bar{F}_{ir/vis}^{i}=f_{loc}^{i}(\bar{F}_{ir/vis}^{i-1}),
\end{split}
\end{equation}
where $\hat{F}_{ir/vis}^{i}$ denotes the intermediate features of the $i$-th block $f_{gl}^{i}$ in the transformer, $\bar{F}_{ir/vis}^{i}$ denotes the intermediate features of the $i$-th block $f_{loc}^{i}$ in the CNN data flow. Besides, when $i=1$, the inputs of $f_{gl}^{i}$ and $f_{loc}^{i}$ (i.e., $\hat{F}_{ir/vis}^{i-1}$ and $\bar{F}_{ir/vis}^{i-1}$) are the source images.

In addition, to perceive the key information of the transformer and the CNN data flow in different scales, we design a feature refinement module, and insert the module into the different depths of the transformer and the CNN data flow to capture the multi-scale global-local features of infrared and visible images. The detailed definition of the refinement module are formulated as follows, i.e.,
\begin{equation}
\begin{split}
\hat{F}_{refine}^{i}=\text{Concat}({{h}_{\max }}(\hat{F}_{ir/vi}^{i}),{{h}_{mean}}(\hat{F}_{ir/vi}^{i})),
\end{split}
\end{equation}
\begin{equation}
\begin{split}
\bar{F}_{refine}^{i}=\text{Concat}({{h}_{\max }}(\bar{F}_{ir/vi}^{i}),{{h}_{mean}}(\bar{F}_{ir/vi}^{i})),
\end{split}
\end{equation}
where $\hat{F}_{refine}^{i}$ and $\bar{F}_{refine}^{i}$ denote the refined features of the transformer and the CNN data flows, respectively. ${{h}_{\max }}(\cdot )$ and ${{h}_{mean}}(\cdot )$ denote the max pooling and the average pooling operations in the channel dimension, and Concat represents the concatenate operation. 

\begin{figure}[!t]
\centering
\includegraphics[width=3.5in]{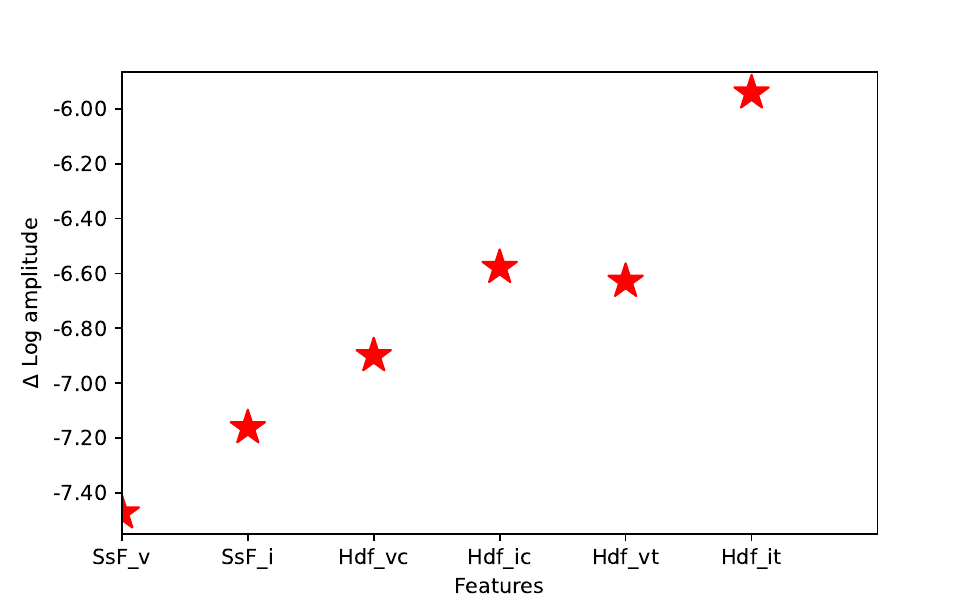}
\caption{The frequency response of different features. The significant semantic features contain more low-frequent information, and the shallow features contain more high-frequent details. The combination of the high-frequent details and low-frequent significant semantic information can improve the visual effect of the fused image.}
\label{FIG:6}
\end{figure}

\subsection{Refined feature adaptive-modulation (RFaM) for semantic segmentation}
In our proposed method, to consider the significance of different features for the fused result, we employ a refined feature adaptive-modulation (RFaM) mechanism in each branch. The architecture of the RFaM is shown in Fig. 4. Specifically, the refined features of the transformer and the CNN data flows ($\hat{F}_{refine}^{i}$ and $\bar{F}_{refine}^{i}$) are first upsampled to the same size, and this is formulated as
\begin{equation}
\begin{split}
F_{refine}^{i}=\text{upsampl}{{\text{e}}^{({{2}^{i}})}}(\text{concat}(\hat{F}_{refine}^{i},\bar{F}_{refine}^{i})),
\end{split}
\end{equation}
where $F_{refine}^{i}$ denotes the combination of the refined features for the transformer and the CNN data flows, $\text{upsampl}{{\text{e}}^{({{2}^{i}})}}$ denotes the upsample operation, and $i=1,2,3,4$ in our model. Meanwhile, we assign every feature a weight to investigate the significance of different features, and the weights are normalized and dynamically updated during the training process, and the weights are summed up to 1. In addition, the weighted features are applied to the multi-layer perceptrons ($\text{MLP}$) to calculate the semantic segmentation result ($\text{Seg}$). The details of the RFaM are formulated as follow, i.e.,
\begin{equation}
\begin{split}
\text{Seg}=\text{MLP}({{w}_{i}}\cdot F_{refine}^{i}),
\end{split}
\end{equation}
where $\sum {{w}_{i}}=1$ in our model. 

\subsection{Pilot experiments to explore significant semantic features}
In our proposed method, to explore the significant semantic information to meet the high-level vision task, we propose a dual-modal semantic guided image fusion method, by designing two parallel semantic segmentation branches. Specifically, we utilize the two branches (i.e., Branch \uppercase\expandafter{\romannumeral1} and Branch \uppercase\expandafter{\romannumeral3} of Fig.~3) to conduct the pilot experiments, which aims to achieve the semantic segmentation tasks for the infrared and visible image, respectively. Moreover, we track and visualize the changes of the weights existing in the RFaM to analyze the significance of every refined feature. Fig.~5 shows the changes of the weights in the infrared and visible image semantic segmentation tasks, which illustrate that each branch relies more on several features with the training progresses. Note that, we will also conduct the ablation experiment to further demonstrate the effectiveness of the significant semantic features in pilot experiments, see details in section~\uppercase\expandafter{\romannumeral4}.

Based on the pilot experiments, we can explore the significant semantic features to guide the fusion task (i.e., Branch \uppercase\expandafter{\romannumeral2} of Fig. 3) in our final training process. Fig.~5 visualizes the features with larger weights, that can capture the significant semantic information. Specifically, we proposed to utilize the significant prior semantics to emphasize the key features and avoid introducing the redundancy information in our proposed model.  

\begin{figure}[!t]
\centering
\includegraphics[width=3.5in]{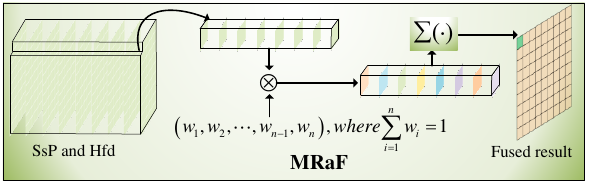}
\caption{The architecture of the multi-level representation-adaptive fusion (MRaF).}
\label{FIG:7}
\end{figure}

\subsection{Multi-level representation-adaptive fusion (MRaF) of significant semantic feature and high-frequent details}

In our proposed method, the fusion model should provide visual appealing fused image and explore the significant semantic information to meet the requirements of high-level vision task. Since the pilot experiments can perceive the significant features, that are semantically distinct enough in each semantic segmentation branch and this can further help us to improve the performance of the proposed method for the downstream task. To achieve the impressive visual effect, we propose to utilize the tool in~\cite{park2022vision} to investigate the frequency response of the significant semantic features. The results of the frequency response are visualized in Fig.~6, which illustrates that the significant semantic features (i.e., $SsF\_i$ and $SsF\_v$) contain more low-frequent information. This observation indicates that the high-frequent information is necessary to be introduced, and may provide more details and improve the visual effect of the fused image. Based on this analysis, we further analyze the frequency response of different features, and find that the shallow features usually contain high-frequent information. Specifically, the frequency response of the shallow features is also shown in Fig.~6, i.e., $Hfd\_ic$, $Hfd\_vc$, $Hfd\_it$ and $Hfd\_it$. Here, $Hfd\_ic$ and $Hfd\_vc$ denote the high-frequent details of the CNN branch for the infrared and visible images, respectively. $Hfd\_it$ and $Hfd\_vt$ denote the high-frequent details of the transformer branch for the infrared and visible images, respectively. 

With the above observation, we proposed to combine the high-frequent details with low-frequency significant semantic information through a multi-level representation-adaptive fusion (MRaF) module. Fig.~7 shows the detailed architecture of the MRaF, and the associated definition of the MRaF is formulated as follows, i.e.,
\begin{equation}
\begin{split}
{{F}_{out}}=\sum{({{w}_{i}}}\cdot SsF+{{w}_{j}}\cdot Hfd),
\end{split}
\end{equation}
where ${{F}_{out}}$ denotes the fused result, $SsF$ and $Hfd$ denote the significant semantic features and the high-frequent details of the shallow features. ${{w}_{i}}$ and ${{w}_{j}}$ are dynamically updated and the weights are summed up to 1 during the training process.

\subsection{Loss function}
In our proposed method, the fused result should aggregate both the high-level semantics and the impressive visual effect. Therefore, the loss function need to simultaneously contain the visual appealing loss and the semantic perception-based loss. The former loss aims to improve the performance of the image visual effect, and the later loss aims to constrain the fused result to capture more semantic information. The loss function of our proposed method is formulated as
\begin{equation}
\begin{split}
{{L}_{total}}={{L}_{visual}}+{{L}_{seg}},
\end{split}
\end{equation}
where ${{L}_{total}}$ denotes the whole loss, ${{L}_{visual}}$ and ${{L}_{seg}}$ denote the visual appealing loss and the semantic perception-based loss, respectively. In ${{L}_{visual}}$, we need to constrain the fused image to preserve more thermal radiation from the infrared image, which is characterized by the intensity information. Therefore, we utilize the intensity loss ${{L}_{\operatorname{int}}}$ to preserve the infrared information. In addition, to capture the details from both infrared and visible images, we also employ the texture loss ${{L}_{tex}}$ \cite{tang2023rethinking}. Thus ${{L}_{visual}}$ can be formulated as
\begin{equation}
\begin{split}
 & {{L}_{\text{visual}}}=\lambda \cdot {{L}_{\operatorname{int}}}+{{L}_{tex}} \\ 
 & =\frac{\lambda }{HW}\cdot \left\| {{F}_{out}}-{{I}_{ir}} \right\|_{F}^{2}+ \\ 
 & \frac{1}{HW}\cdot {{\left\| \left| \nabla {{F}_{out}} \right|-\max \left( \left| \nabla {{I}_{ir}} \right|,\left| \nabla {{I}_{vi}} \right| \right) \right\|}_{1}},  
\end{split}
\end{equation}
where ${{F}_{out}}$ denotes the fused result, $\nabla $ denotes the Sobel gradient operator, ${{\left\| \cdot  \right\|}_{1}}$ represents the ${{l}_{1}}$-norm, ${{\left\| \cdot  \right\|}_{F}}$ is the matrix Forbenius norm, and $\left| \cdot  \right|$ denotes the absolute operation. For the semantic perception-based loss ${{L}_{seg}}$, we utilize the OHEMCELoss \cite{shrivastava2016training} to preserve the semantic information. 

\section{Experiments}
In this section, we compare the performance of the proposed method with several existing alternative methods in terms of the semantic segmentation. Specifically, we adopt the excellent pretrained semantic segmentation models associated with their original parameters to conduct the segmentation experiments on two public datasets. In addition, to demonstrate the superiority of the proposed method in visual effects, we also compare the proposed method with the other alternative methods in both qualitative and quantitative ways. Finally, we conduct several ablation experiments to demonstrate the rationality of the proposed model, and provide the efficiency comparisons for all methods under comparisons.


\begin{figure}[!t]
\centering
\includegraphics[width=3.5in]{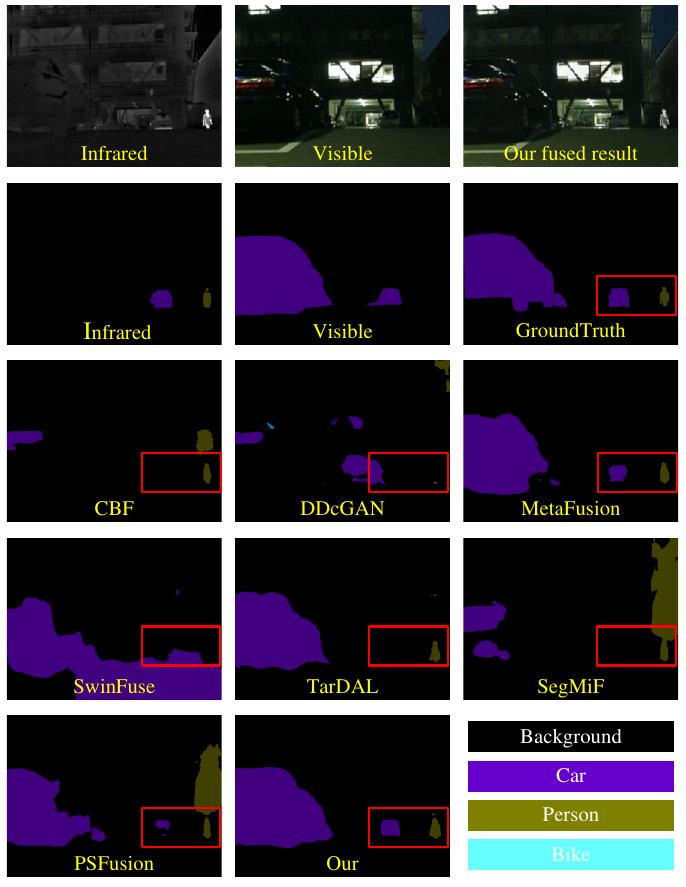}
\caption{An example of semantic segmentation result of different methods on MSRS dataset by Segformer pretrained model.}
\label{FIG:8}
\end{figure}

\begin{figure}[!t]
\centering
\includegraphics[width=3.5in]{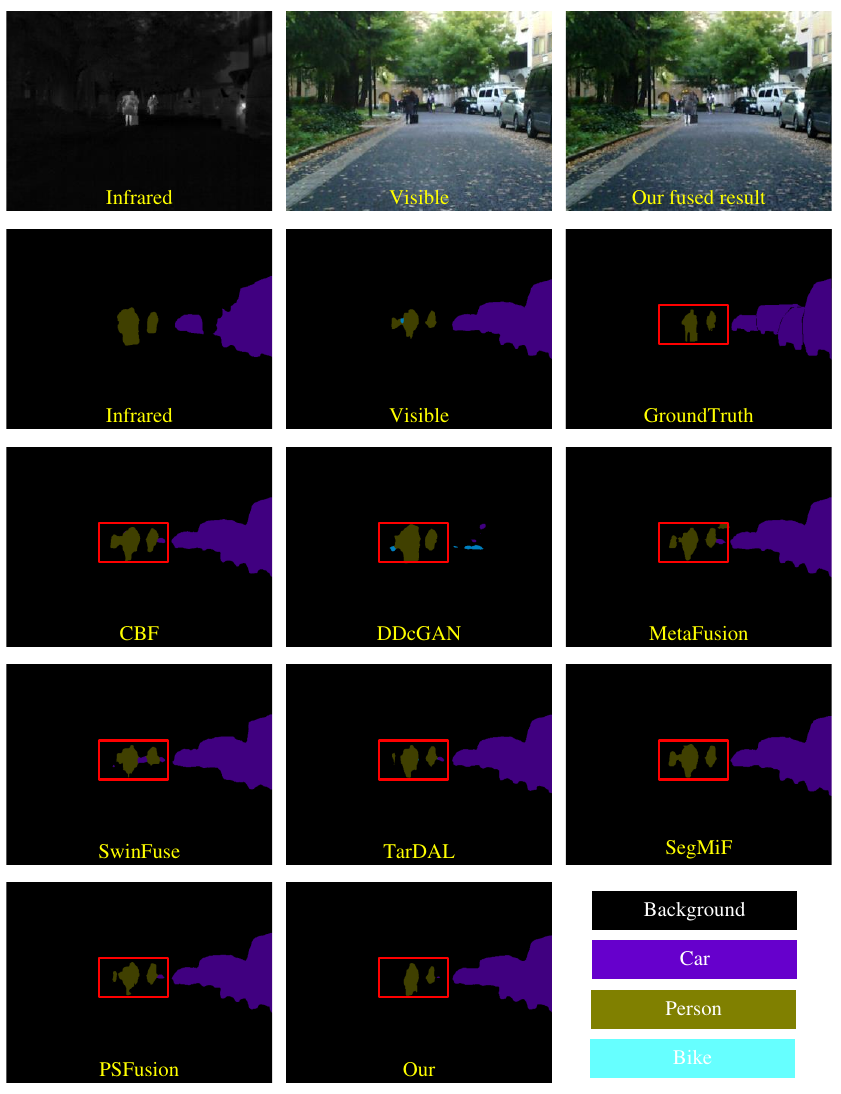}
\caption{Another example of semantic segmentation result of different methods on MSRS dataset by ViT-Adapter pretrained model.}
\label{FIG:9}
\end{figure}

\subsection{Experimental configurations}
For the proposed method, we aim to fuse the infrared and visible images to achieve the good performance both in the high-level vision task and the impressive visual effect. To demonstrate the effectiveness of the proposed method, we compare it with other existing alternative methods in both semantic segmentation and fusion tasks. We employ the MSRS \cite{tang2023rethinking} and FMB~\cite{liu2023SegMiF} datasets for the experiments, and the reason of utilizing the two dataset is that they contain aligned image pairs and annotated segmentation labels. Specifically, we first compare the proposed method with other alternative methods on the MSRS dataset for the semantic segmentation and fusion tasks. Then we analyze the generalization of the proposed method, and compare it with other alternative methods on the FMB dataset. Specifically, we compare the proposed method with several high-level vision task-driven methods (i.e., the PSFusion~\cite{tang2023rethinking}, the SegMiF~\cite{liu2023SegMiF}, the TarDAL~\cite{liu2022target}), which include either the semantic segmentation task-driven or the object detection task-driven methods. Finally, we compare the proposed method with several vision-perception oriented methods (i.e., the CBF~\cite{shreyamsha2015image}, the DDcGAN~\cite{ma2020ddcgan}, the MetaFusion~\cite{li2021different}, the SwinFuse~\cite{wang2022swinfuse}), which include the traditional methods, CNN-based methods, meta learning-based methods and Transformer-based methods.

For the semantic segmentation task, to objectively evaluate the performance of the proposed method and other methods, we propose to employ the representative pretrained semantic segmentation models associated with their original parameters, e.g., the Segformer~\cite{xie2021segformer}, the ViTadpter~\cite{chen2023vitadapter}. For the image fusion task, we compare the proposed method with other alternative methods in both qualitative and quantitative ways. For the qualitative experiments, we evaluate the performance of all methods based on the human visual inspection, i.e., the illuminance, the sharpness, the contrast. For the quantitative experiments, we adopt four evaluation metrics, including the mutual information (MI)~\cite{qu2002information}, the structural similarity index measure (SSIM)~\cite{wang2004image}, the peak signal-to-noise ratio (PSNR) as well as the sum of correlation (SCD)~\cite{aslantas2015new}, and the larger values of the above metrics with better performance. 

\subsection{Implementation details}
In our experiments, we train the proposed model on the MSRS dataset, and we also take the data augmentation technique during the training process. For example, we crop the training image into the sub-image with the size of 256×256, and the stride of the cropping is set as 100. In addition, we adopt the Adam optimizer to update the parameters, and the learning rate is
initialized as $1\times {{10}^{\text{-}4}}$. Specifically, the batch size is set as 20. Besides, we set the hyper-parameters $\lambda$ as 0.1, and we train the proposed model through the NVIDIA GeForce RTX 4090 with 24 GB memory.

\begin{figure}[!t]
\centering
\includegraphics[width=3.5in]{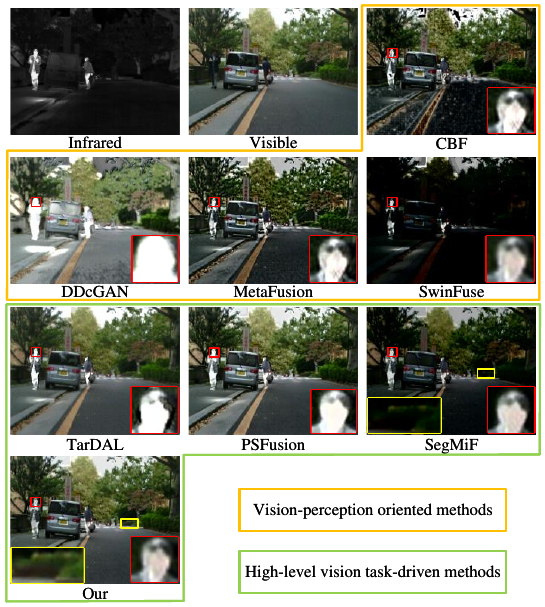}
\caption{An example of fused result of different methods on MSRS dataset.}
\label{FIG:10}
\end{figure}

\subsection{Experimental Comparisons on the MSRS Dataset}
To demonstrate the superiorities of the proposed method both in the high-level vision task and the impressive visual effect, we compare the proposed method with other alternative methods on the MSRS dataset both in the semantic segmentation task and the image fusion task.

\subsubsection{Comparisons and analysis in the semantic segmentation task}
For the semantic segmentation task, we compare our fused result with other alternative methods based on two pretrained segmentation models (i.e., the Segformer and the ViT-Adapter). To objectively evaluate the performance of different fusion methods, we take their original parameters of the pretrained models without fine tuning. Specifically, Fig. 8 shows the visualization segmentation results based on the Segformer. The first row of Fig. 8 denotes the source images and the fused result of the proposed method, and the second row shows the segmentation result of the infrared and visible images, as well as the ground truth. Note that, the segmentation result of the single modal image, such as the infrared or visible image, fails to capture all the targets. For example, the segmentation result of the infrared image cannot classify the vehicle existing in the left of the image. On the contrary, the result of the visible image omits the person existing in the right of the image. However, our result can classify all the vehicles and persons, this is because we refine and explore the infrared and visible features that are semantically distinct enough to guide the fusion model, which improves the quality of learned representations for fusion results in the high-level vision task.

For Fig.8, compared with the ground truth, only our result and the TarDAL can segment the vehicle existing in the left of the scene, however our segmentation result has better contour than the TarDAL. In addition, the regions in the red blocks of all methods also show that only the proposed method and the MetaFusion can produce correct segmentation result as the ground truth, but our result has more accurate classification with the precise contour for the vehicle. 

On the other hand, Fig. 9 shows the visualization segmentation results based on the ViT-Adapter, which also illustrates that our fused result can provide more accurate semantics than the single infrared or visible image. Meanwhile, for the visualized segmentation results of the alternative methods, most of the fused methods can correctly segment the vehicles in the scene, except the DDcGAN method. However, the regions existing in the red block of all the methods show that only our result can correctly segment the pedestrians without introducing other incorrect labels. By contrast, the alternative methods contain other incorrect classifications for the pixels in the red blocks. 

\begin{figure}[!t]
\centering
\includegraphics[width=3.5in]{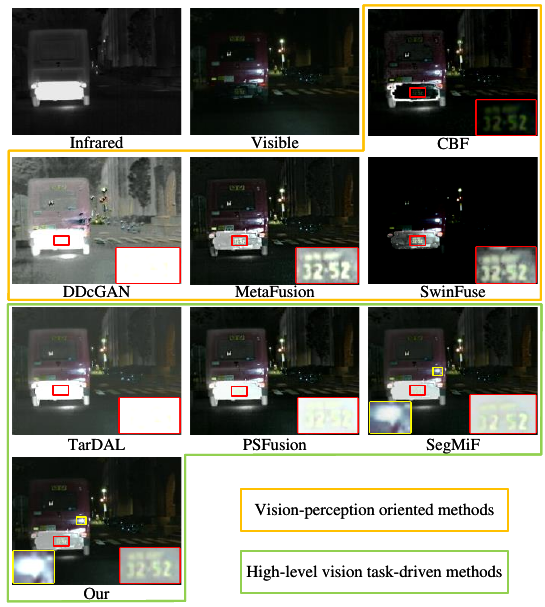}
\caption{Another example of the fused result based on the different methods on the MSRS dataset.}
\label{FIG:11}
\end{figure}

\begin{figure}[!h]
\centering
\includegraphics[width=3.5in]{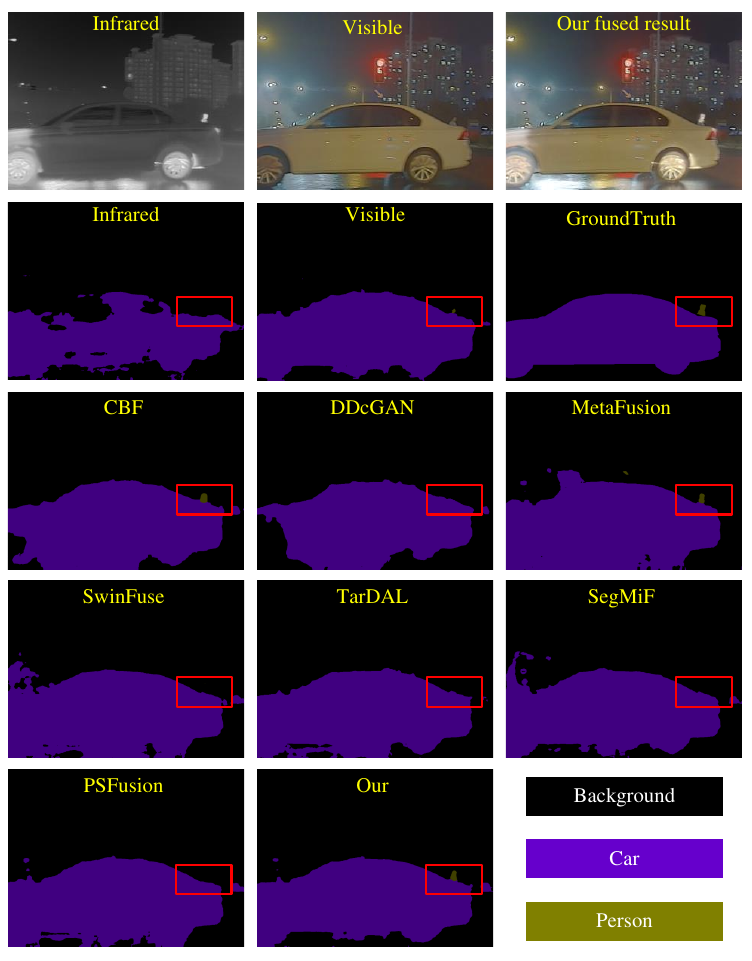}
\caption{An example of the semantic segmentation result based on different methods on the FMB dataset with the Segformer pretrained model.}
\label{FIG:12}
\end{figure}

Furthermore, we also compare the proposed method with other alternative methods in the high-level vision task with the quantitative way. Table \uppercase\expandafter{\romannumeral1} reports the quantitative analysis results of different categories on the MSRS dataset, which illustrate that the proposed method has better performance than other alternative methods for all the categories. Note that, the proposed method not only achieves the highest IoU for different categories in most cases, but also has the highest mean IoU (mIoU) among all the compared methods. The qualitative and quantitative analysis demonstrates the superiorities of the proposed method.

\subsubsection{The Comparison and analysis for the fusion task}
To demonstrate the advantages of the proposed method in the fusion task, we also compare it with other alternative methods with both in qualitative and quantitative ways. For the qualitative analysis, Fig. 10 and Fig. 11 show the fused results of all the methods, which include the vision-perception oriented methods and the high-level vision task-driven methods. For Fig. 10, we select the representative regions and zoom them in the bottom of the images, which show that the MetaFusion, the SwinFuse, the SegMiF and the proposed method preserve more texture information in the red blocks. However, the alternative MetaFusion, SwinFuse and SegMiF all fail to present the satisfactory background information. For example, the region existing in yellow block of the proposed method contains more details than the SegMiF.

For Fig. 11, the proposed method also has better visual effect than other alternative methods. For example, only the MetaFusion, the SwinFuse, the SegMiF and the proposed method can clearly present the numbers that existing in the red block, but the MetaFusion and the SwinFuse fail to preserve enough background information and contain low contrast. On the contrary, the SegMiF and the proposed method not only clearly present the numbers, but also preserve more background information. However, the SegMiF still fails to capture the details of the visible image, such as the car lamp based on the proposed method is clearer in the yellow block than the SegMiF.

For the quantitative analysis, we adopt the multiple metrics to objectively evaluate all the methods, which include the MI, the SSIM, the PSNR and the SCD. Table \uppercase\expandafter{\romannumeral2} shows that the proposed method achieves the better performance than other alternative methods on the MI, the SSIM and the PSNR, and it also has acceptable performance on the SCD (ranked as the second). Specifically, the proposed method has the largest MI value, which illustrates that our method captures more meaningful information from the source images. The largest SSIM and PSNR values of our method also demonstrate that our result can preserve more structure information and introduce little noise from the source images. For the SCD, the proposed method only follows the PSFusion method with a narrow margin, but our method still has the better visual effect than the PSFusion, such as the examples in Fig. 10 and Fig. 11. 

\begin{table}[]
\centering
\caption{{Quantitative analysis results of the semantic segmentation for different methods on the MSRS dataset. The \textbf{bold} values indicate the best model performance, and the \textcolor{red}{red} values denote the second order.}}
\setlength{\tabcolsep}{0.35 mm}
\renewcommand\arraystretch{1.1}{
\begin{tabular}{lccccc|ccccc}
\hline
\multicolumn{1}{c}{} & \begin{tabular}[c]{@{}c@{}}Backg-\\ round\end{tabular} & \multicolumn{1}{l}{\begin{tabular}[c]{@{}l@{}}Pedes-\\ trian\end{tabular}} & Car                          & Bike                         & mIoU                         & \begin{tabular}[c]{@{}c@{}}Backg-\\ round\end{tabular} & \multicolumn{1}{l}{\begin{tabular}[c]{@{}l@{}}Pedes-\\ trian\end{tabular}} & Car                          & Bike                         & mIoU                         \\ \hline
Backbone             & \multicolumn{5}{c|}{ViT-Adapter}                                                                                                                                                                                                 & \multicolumn{5}{c}{Segformer}                                                                                                                                                                                                    \\ \hline
Visible              & 95.87                                                  & 63.43                                                                      & 54.86                        & 55.25                        & 67.35                        & 95.77                                                  & 62.27                                                                      & 53.18                        & 50.90                        & 65.53                        \\
Infrared             & 94.41                                                  & 37.72                                                                      & 53.66                        & 10.57                        & 49.09                        & 94.42                                                  & 39.48                                                                      & 49.83                        & 18.65                        & 50.59                        \\ \hline
CBF                  & 95.54                                                  & 57.63                                                                      & 54.29                        & 43.96                        & 62.85                        & 95.58                                                  & 60.59                                                                      & 55.07                        & 41.48                        & 63.18                        \\
DDcGAN               & 94.64                                                  & 49.27                                                                      & 38.87                        & 15.78                        & 49.64                        & 94.76                                                  & 51.62                                                                      & 38.06                        & 23.78                        & 52.06                        \\
MetaFusion           & 95.83                                                  & 61.50                                                                      & 58.47                        & 50.38                        & 66.55                        & {\color[HTML]{FF0000} 95.87}                           & \textbf{63.62}                                                             & 60.02                        & 46.31                        & 66.46                        \\
SwinFuse             & 94.62                                                  & 52.19                                                                      & 34.15                        & 28.04                        & 52.25                        & 93.63                                                  & 40.91                                                                      & 39.05                        & 27.15                        & 50.19                        \\
TarDAL               & 95.89                                                  & 62.70                                                                      & 61.39                        & 47.77                        & 66.94                        & \textbf{95.89}                                         & 63.25                                                                      & {\color[HTML]{FF0000} 60.88} & 47.29                        & 66.83                        \\
PSFusion             & {\color[HTML]{FF0000} 95.94}                           & {\color[HTML]{FF0000} 63.71}                                               & {\color[HTML]{FF0000} 61.62} & {\color[HTML]{FF0000} 51.25} & {\color[HTML]{FF0000} 68.13} & 95.85                                                  & 62.73                                                                      & \textbf{61.02}               & {\color[HTML]{FF0000} 49.59} & {\color[HTML]{FF0000} 67.30} \\
SegMiF               & 95.82                                                  & 62.62                                                                      & 58.85                        & 48.04                        & 66.33                        & 95.76                                                  & 61.99                                                                      & 58.52                        & 45.60                        & 65.47                        \\
Our                  & \textbf{96.01}                                         & \textbf{64.61}                                                             & \textbf{61.63}               & \textbf{53.19}               & \textbf{68.86}               & \textbf{95.89}                                         & {\color[HTML]{FF0000} 63.47}                                               & 60.26                        & \textbf{50.38}               & \textbf{67.50}               \\ \hline
\end{tabular}}
\end{table}
\subsection{Generalization experiments on the FMB dataset}
In our experiments, we have compared the proposed method with other alternative methods on the MSRS dataset to demonstrate the effectiveness and superiorities. To further illustrate the ability of the generalization for our method, we also evaluate the performance of all the methods on the FMB dataset. The generalization experiments on the FMB dataset include the comparison and analysis for the semantic segmentation task and the fusion task.

\begin{figure}[!h]
\centering
\includegraphics[width=3.5in]{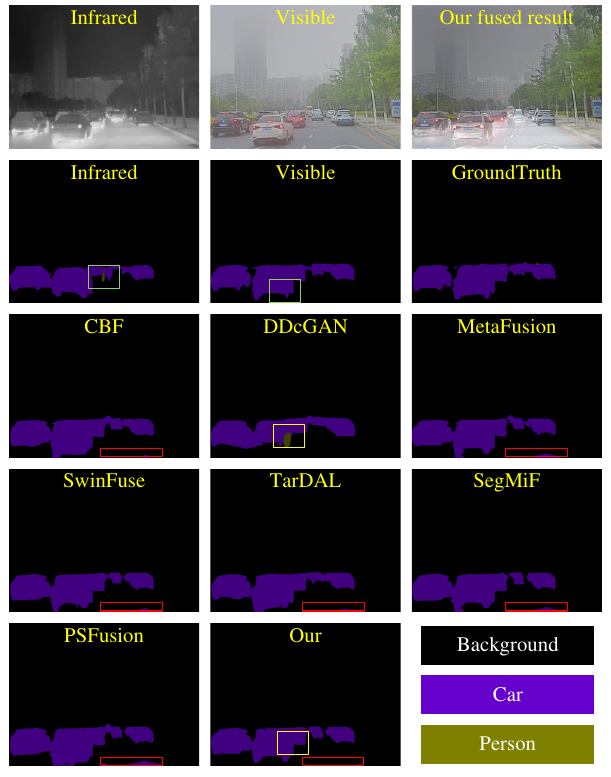}
\caption{Another example of semantic segmentation result of different methods on FMB dataset by ViT-Adapter pretrained model.}
\label{FIG:13}
\end{figure}

\begin{figure}[!h]
\centering
\includegraphics[width=3.5in]{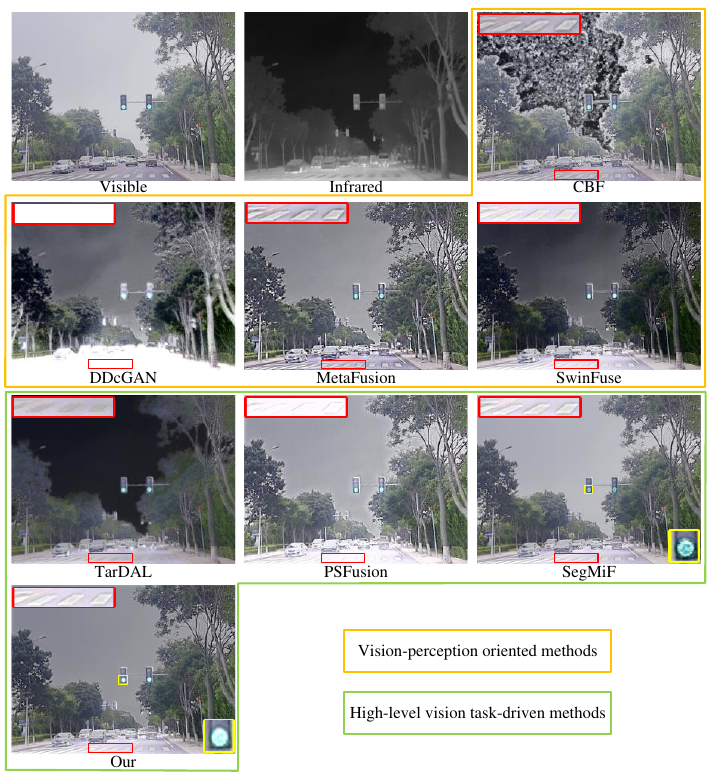}
\caption{An example of fused result of different methods on FMB dataset.}
\label{FIG:14}
\end{figure}

\subsubsection{The comparison and analysis for the semantic segmentation task}
Similar to the former experiments in the semantic segmentation task, we also adopt the pretrained segmentation models (i.e., the Segformer and the ViT-Adapter) to objectively evaluate the performance of all the methods with their original parameters. Fig. 12 shows the visualization segmentation results based on the Segformer, and the first row shows the source images and the fused result of the proposed method. The second row presents the segmentation results of the source images and the ground truth. Note that, the segmentation result of our method in the last row has the better performance than the single infrared or visible image. For example, the segmentation result of the infrared images fails to classify the pedestrian existing in the red block, and has the incomplete contour of the vehicle. Besides, the result of the visible image contains less details of the pedestrian in the red block than ours. Compared with other alternative methods, only the CBF, the MetaFusion and the proposed method can correctly segment the pedestrian in the red block, but only our segmentation result has more details and the completed contour profile than the CBF and the MetaFusion.

In addition, Fig. 13 shows the visualization segmentation results based on the ViT-Adapter, and our segmentation result also has the better performance than the single infrared or visible image. For example, the segmentation result in the green block of the infrared image contains incorrect classification results, and the segmentation result in the green block of the visible image fails to capture the complete contour of the vehicle. Moreover, most segmentation results based on the alternative methods contain incorrectly classification pixels, such as the region existing in red blocks of the CBF, the MetaFusion, the SwinFuse, the TarDAL, the SegMiF and the PSFusion. Meanwhile, the segmentation results in the yellow block of the TarDAL fails to correctly segment the vehicle. By contrast, the proposed method has the better segmentation result than all the alternative methods.  

Besides, we also evaluate the performance of all the methods for semantic segmentation task through the quantitative way. Table \uppercase\expandafter{\romannumeral3} shows the quantitative analysis results of different categories on the FMB dataset. The quantitative analysis results illustrate that our method not only has the highest IoU for all different categories, but also has the highest mean mIoU than all the alternative methods. As a result, both the qualitative and the quantitative analysis of the generalization experiments demonstrate that the proposed method has the better performance than both the vision-perception oriented methods and the high-level vision task-driven methods.
\begin{table*}[]
\centering
\caption{{Quantitative analysis results of image fusion for different methods on the MSRS and FMB datasets. The \textbf{bold} values indicate the best model performance, and the \textcolor{red}{red} and \textcolor{blue}{blue} values denote the second and third order.}}
\setlength{\tabcolsep}{1.5 mm}
\renewcommand\arraystretch{1.1}{
\begin{tabular}{c|cccc|cccc|ccc}
\hline
\textbf{}  & MI                          & SSIM                        & PSNR                         & SCD                         & MI                          & SSIM                        & PSNR                         & SCD                         & \multicolumn{3}{c}{Efficient Analysis}       \\ \hline
           & \multicolumn{4}{c|}{MSRS}                                                                                              & \multicolumn{4}{c|}{FMB}                                                                                               & Parameters(M) & Time(s)-MSRS & Time(s)-FMB   \\ \hline
CBF        & 2.35                        & {\color[HTML]{00B0F0} 1.10} & {\color[HTML]{FF0000} 16.47} & 1.12                        & 2.05                        & 0.99                        & {\color[HTML]{00B0F0} 14.86} & 0.93                        & -             & -            & -             \\
DDcGAN     & 1.65                        & 0.55                        & 8.05                         & 1.01                        & 2.40                        & 1.07                        & 11.44                        & 1.14                        & 1.097         & 0.53         & 0.56          \\
MetaFusion & 1.45                        & 1.08                        & 15.86                        & 1.36                        & 2.15                        & 1.18                        & 14.37                        & 1.39                        & 2.15          & 0.26         & 0.27          \\
SwinFuse   & 1.78                        & 0.62                        & 15.33                        & 1.03                        & {\color[HTML]{FF0000} 2.81} & 1.20                        & 13.53                        & \textbf{1.66}               & 23.07         & 0.48         & 0.58          \\
TarDAL     & {\color[HTML]{00B0F0} 2.60} & 1.00                        & 13.58                        & 1.43                        & \textbf{2.87}               & {\color[HTML]{FF0000} 1.33} & {\color[HTML]{FF0000} 14.95} & 0.91                        & \textbf{0.28} & 0.42         & 0.51          \\
PSFusion   & {\color[HTML]{FF0000} 2.83} & {\color[HTML]{FF0000} 1.16} & {\color[HTML]{00B0F0} 16.15} & \textbf{1.70}               & 2.50                        & 1.24                        & 12.95                        & {\color[HTML]{FF0000} 1.63} & 43.83         & 0.21         & 0.27          \\
SegMiF     & 2.41                        & \textbf{1.19}               & 16.12                        & {\color[HTML]{00B0F0} 1.46} & {\color[HTML]{00B0F0} 2.71} & {\color[HTML]{00B0F0} 1.32} & {\color[HTML]{FF0000} 14.95} & 1.45                        & 43.53         & 0.82         & 0.97          \\
Our        & \textbf{2.85}               & \textbf{1.19}               & \textbf{16.98}               & {\color[HTML]{FF0000} 1.60} & 2.54                        & \textbf{1.36}               & \textbf{15.05}               & {\color[HTML]{00B0F0} 1.58} & 10.32         & \textbf{0.1} & \textbf{0.11} \\ \hline
\end{tabular}}
\end{table*}

\subsubsection{The comparison and analysis for the fusion task}
For the generalization experiments, we also compare the proposed method with other alternative fusion methods for the image fusion task, and we evaluate our fused result with others by both qualitative and quantitative ways. Fig. 14 and Fig. 15 show the fused results of different methods, which include the vision-perception oriented methods and the high-level vision task-driven methods. For Fig. 14, only the proposed method and the SegMiF can preserve more texture information than other methods in the red blocks that we zoom in the top of the fused image. Meanwhile, our method contains more details than the SegMiF, such as the traffic light existing in the yellow block is clearer with our method. In Fig. 15, we also select representative regions from each fused result and zoom them in the bottom of the image. We find that expect for our method and the SegMiF, the other methods all contain halo around the lamp, which leads to an unnatural visual effect. Overall, our method has the  better visual effect than all other alternative methods. 

\begin{figure}[!h]
\centering
\includegraphics[width=3.5in]{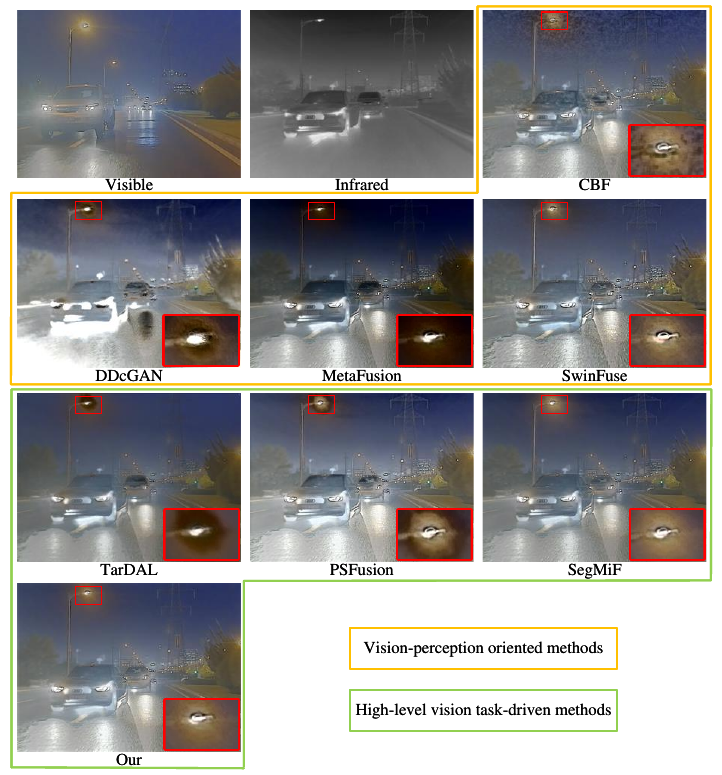}
\caption{Another example of fused result of different methods on FMB dataset.}
\label{FIG:15}
\end{figure}

\begin{figure}[!h]
\centering
\includegraphics[width=3.5in]{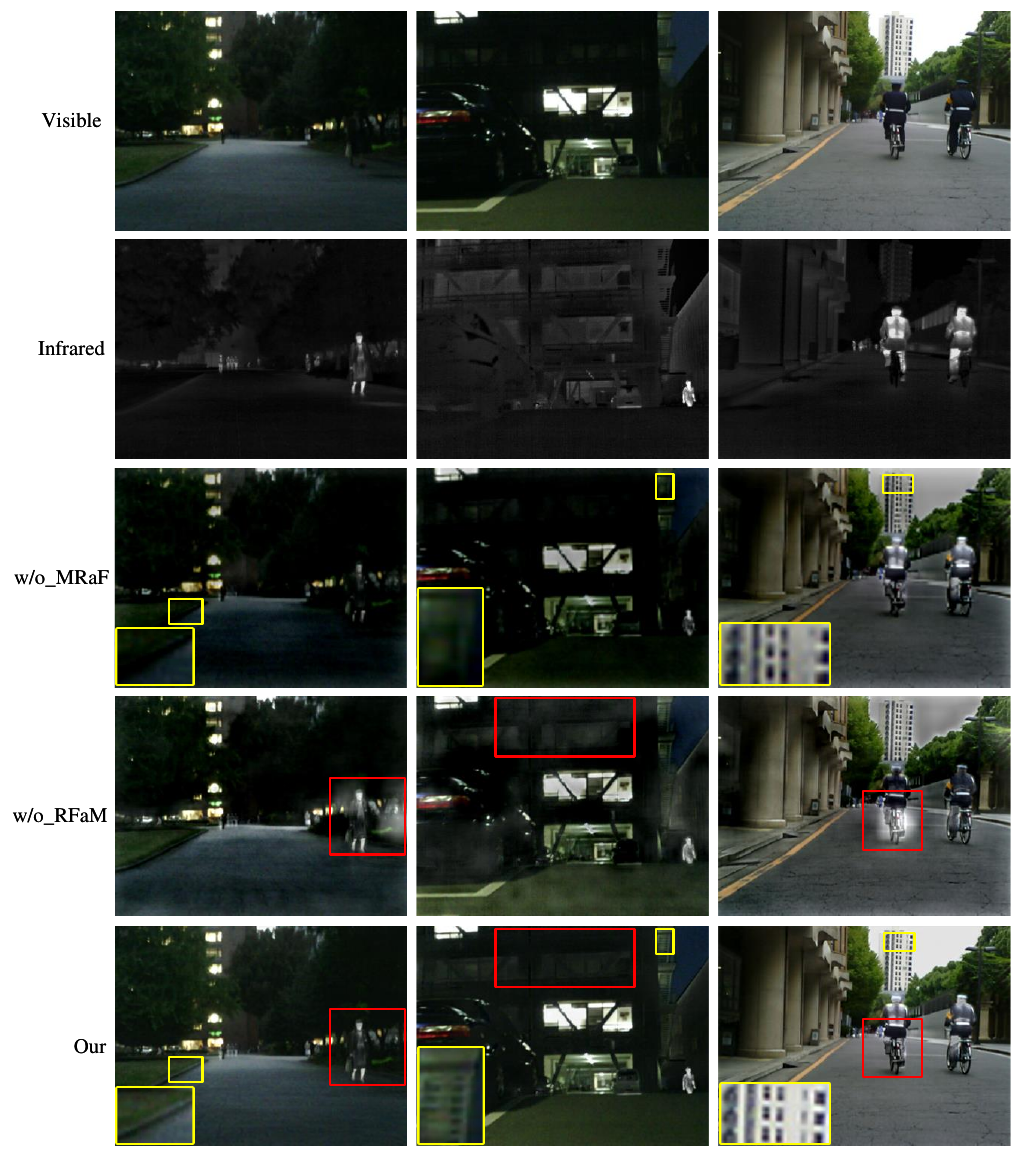}
\caption{Ablation experiments of our method. w/o\_RFaM aims to demonstrate the effectiveness of the significant semantic prior n our methods, and w/o\_MRaF can illustrate the necessity of the combination of high-frequency details and the low-frequency significant semantic feature.}
\label{FIG:16}
\end{figure}

In addition, we also compare our method with other methods in quantitative ways. Table \uppercase\expandafter{\romannumeral2} shows the quantitative analysis results of all the methods, which illustrates that our method achieves the best performance based on the SSIM and PSNR metrics in the generalization experiments, and it also has acceptable performance based on SCD (ranked as third). Our method fails to get the best result based on the MI, but it still has better performance than the PSFusion, the MetaFusion, the DDcGAN and the CBF. Therefore, for the image fusion task of the generalization experiments, our method has better performance both in the qualitative and the quantitative analysis experiments.

\begin{table}[]
\centering
\caption{{Quantitative analysis results of semantic segmentation for different methods on the FMB dataset. The \textbf{bold} values indicate the best model performance, and the \textcolor{red}{red} values denote the second order.}}
\setlength{\tabcolsep}{0.7 mm}
\renewcommand\arraystretch{1.1}{
\begin{tabular}{lcccc|cccc}
\hline
           & \begin{tabular}[c]{@{}c@{}}Backg-\\ round\end{tabular} & \begin{tabular}[c]{@{}c@{}}Pedes-\\ trian\end{tabular} & Car                          & mIoU                         & \begin{tabular}[c]{@{}c@{}}Backg-\\ round\end{tabular} & \begin{tabular}[c]{@{}c@{}}Pedes-\\ trian\end{tabular} & Car                          & mIoU                         \\ \hline
Backbone   & \multicolumn{4}{c|}{ViT-Adapter}                                                                                                                                              & \multicolumn{4}{c}{Segformer}                                                                                                                                                 \\ \hline
Visible    & 97.33                                                  & 51.14                                                  & 66.54                        & 71.67                        & 97.07                                                  & 42.94                                                  & 65.93                        & 68.65                        \\
Infrared   & 97.03                                                  & 50.01                                                  & 62.06                        & 69.70                        & 96.50                                                  & 38.19                                                  & 60.76                        & 65.15                        \\ \hline
CBF        & 97.30                                                  & 34.06                                                  & 68.82                        & 66.73                        & 97.10                                                  & 34.99                                                  & 66.15                        & 66.08                        \\
DDcGAN     & 97.17                                                  & 35.78                                                  & 66.49                        & 66.48                        & 96.90                                                  & 38.39                                                  & 62.91                        & 66.06                        \\
MetaFusion & 97.36                                                  & 38.73                                                  & 68.97                        & 68.35                        & 97.20                                                  & 43.48                                                  & 66.42                        & 69.03                        \\
SwinFuse   & 97.37                                                  & 39.54                                                  & 69.02                        & 68.65                        & 97.19                                                  & 44.10                                                  & 66.14                        & 69.14                        \\
TarDAL     & 97.37                                                  & {\color[HTML]{FE0000} 39.73}                           & 69.07                        & {\color[HTML]{FF0000} 68.72} & 97.20                                                  & {\color[HTML]{FF0000} 44.70}                           & 66.31                        & {\color[HTML]{FF0000} 69.40} \\
PSFusion   & {\color[HTML]{FE0000} 97.38}                           & 38.94                                                  & {\color[HTML]{FE0000} 69.21} & 68.51                        & {\color[HTML]{FF0000} 97.22}                           & 43.90                                                  & {\color[HTML]{FF0000} 66.66} & 69.26                        \\
SegMiF     & 97.37                                                  & 39.65                                                  & 68.94                        & 68.65                        & 97.18                                                  & 44.16                                                  & 65.99                        & 69.11                        \\
Our        & \textbf{97.45}                                         & \textbf{44.20}                                         & \textbf{70.07}               & \textbf{70.58}               & \textbf{97.35}                                         & \textbf{48.17}                                         & \textbf{68.20}               & \textbf{71.24}               \\ \hline
\end{tabular}}
\end{table}

\subsection{Ablation study}
In our method, we analysis the weights of RFaM in pilot experiments to find the features that are semantically distinct enough in semantic segmentation task. Therefore, to demonstrate the effectiveness of the significant semantic prior in our methods, we conduct an ablation experiment by dropping RFaM, which is named w/o\_RFaM. The last two rows of Fig. 16 show several examples of our method and w/o\_RFaM, which illustrates that the results of w/o\_RFaM contain more redundancy information than ours, such as the regions in the red blocks, because our method adopt RFaM in the pilot experiments to capture significant semantic prior, which can reduce redundancy information in the fused result. 

In addition, to achieve impressive visual effect in our fused result, we investigate the frequency response of the significant semantic features, and introduce the high-frequency details into the low-frequency significant semantic feature by a multi-level representation-adaptive fusion (MRaF) module. Thus, we adopt an ablation experiment by dropping high-frequency details, which named w/o\_MRaF. Noting that, our method can preserve more details than w/o\_MRaF. For example, in Fig. 16, the regions in the yellow blocks of our results contains clearer texture and details, which demonstrates the necessary and effectiveness of the combinations of high-frequency details and low-frequency significant semantic feature in MRaF.

\subsection{Efficiency analysis}
In our work, we also compare the efficiency of our method, the vision-perception oriented methods and the high-level vision task-driven methods. We first calculate the parameter size of all the methods, and the results are shown in Table \uppercase\expandafter{\romannumeral2}, which illustrate that our method contains more parameters than the DDcGAN, the MetaFusion and the TarDAL, because our method employes the dual semantic segmentation branches to extract the independent semantics of each modality. However, our method still has less parameters than the SwinFuse, the PSFusion and the SegMiF. Note that, the proposed method, the PSFusion and the SegMiF all take the semantic segmentation as the high-level vision task, but our method uses less parameters, because we propose the refined feature adaptive-modulation (RFaM) module and employ the pilot experiments to extract the significant semantic features rather than all the features. In addition, we calculate the average running time of each method on the two datasets to compare their efficiency. Table \uppercase\expandafter{\romannumeral2} shows that our method consumes less time than the alternative methods, which illustrates that our method has the best computational efficiency.

\section{Conclusion}
In this work, to provide the visual appealing fused image that can explore the significant semantic information to meet the high-level vision task, we have proposed a dual-modality semantic guided image fusion method for high-level vision tasks, that can perceive the significant semantic features and improve the performance of the fusion results in ITS. Specifically, we have designed two parallel semantic segmentation branches to extract the independent semantics of each modality. Moreover, we have analyzed the weights of the RFaM in the pilot experiments to perceive the features that are semantically distinct enough in each semantic segmentation branch, i.e., the features can be seen as the significant prior sematics to guide the image fusion task. To achieve the impressive visual effect for our fused result, we have investigated the frequency response of the significant semantic features, and proposed the MRaF module to explicitly integrate the low-frequent significant semantic features with the high-frequent details. Extensive experiments and ablation analysis have demonstrated the effectiveness and superiority of our methods over the fusion approaches both in the visual appeal and the high-level vision task.

\ifCLASSOPTIONcaptionsoff
  \newpage
\fi

\end{document}